\documentclass[preprint,12pt]{elsarticle}

\usepackage{amssymb}
\usepackage{amsmath}
\usepackage{float}
\usepackage{color}
\usepackage[ruled,vlined]{algorithm2e}

\journal{European Journal of Control}

\begin{document}

\begin{frontmatter}



\title{
On Using Machine Learning to Early Detect Catastrophic Failures in Marine Diesel Engines
}


\author[label1]{Francesco Maione} 
\author[label1]{Paolo Lino} 
\author[label2]{Giuseppe Giannino}
\author[label1]{Guido Maione} 

\affiliation[label1]{organization={Politecnico di Bari},
            addressline={Via E. Orabona, 4}, 
            city={Bari},
            postcode={70125}, 
            country={Italy}}
\affiliation[label2]{organization={Isotta Fraschini Motori},
            addressline={V.le Francesco de Blasio, 1}, 
            city={Bari},
            postcode={70132}, 
            country={Italy}
            }

\begin{abstract}

{

Catastrophic failures of marine engines imply severe loss of functionality and destroy or damage the systems irreversibly.
Being sudden and often unpredictable events, they pose a severe threat to navigation, crew, and passengers. The abrupt nature makes early detection the only effective countermeasure. However, research has concentrated on modeling the gradual degradation of components, with limited attention to sudden and anomalous phenomena. 
This work proposes a new method for early detection of catastrophic failures. Based on real data from a failed engine, 
the approach evaluates the derivatives of the deviation between actual sensor readings and expected values of engine variables. Predictions are obtained by a Random Forest, which is the most suitable Machine Learning algorithm among the tested ones. 
Traditional methods focus on deviations of monitored signals, whereas the proposed approach employs the derivatives of the deviations to provide earlier indications of abnormal dynamics, and to alert that a rapid and dangerous event is breaking out 
within the system. The method allows the detection of anomalies before measurements reach critical thresholds and alarms are triggered, 
which is the common method in industry. 
Consequently, operators can be warned in advance and shut down the engine, then prevent 
damage and unexpected power loss. Moreover, they have the time to safely change the ship route {and avoid} potential obstacles. 
Simulation results confirm the effectiveness of the proposed approach 
in anticipating occurrence of catastrophic failures. Validation on real-world data further reinforces the robustness and practical applicability of the method. It is worth noting that data acquisition to train the predictive algorithm is not a problem, since a Deep Learning-based data augmentation procedure is used.

}

\end{abstract}

\begin{graphicalabstract}
\includegraphics[width=\linewidth, height=7.5cm]{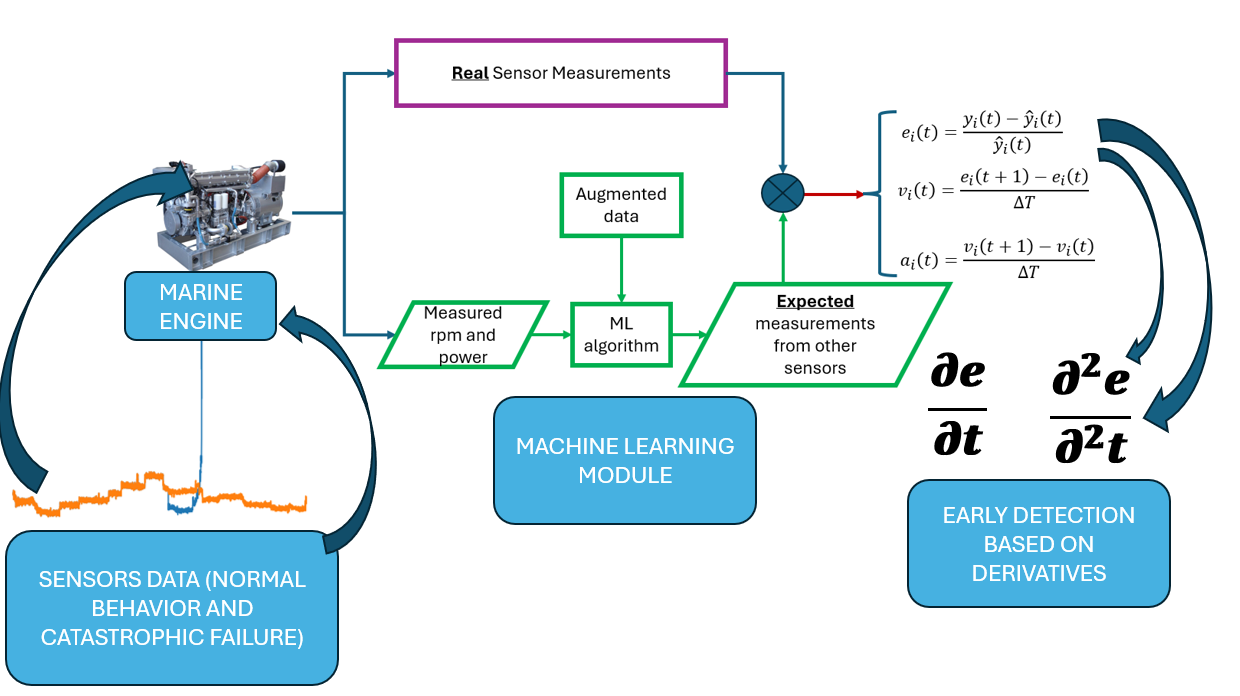} \\
\includegraphics[width=\linewidth, height=3.5cm]{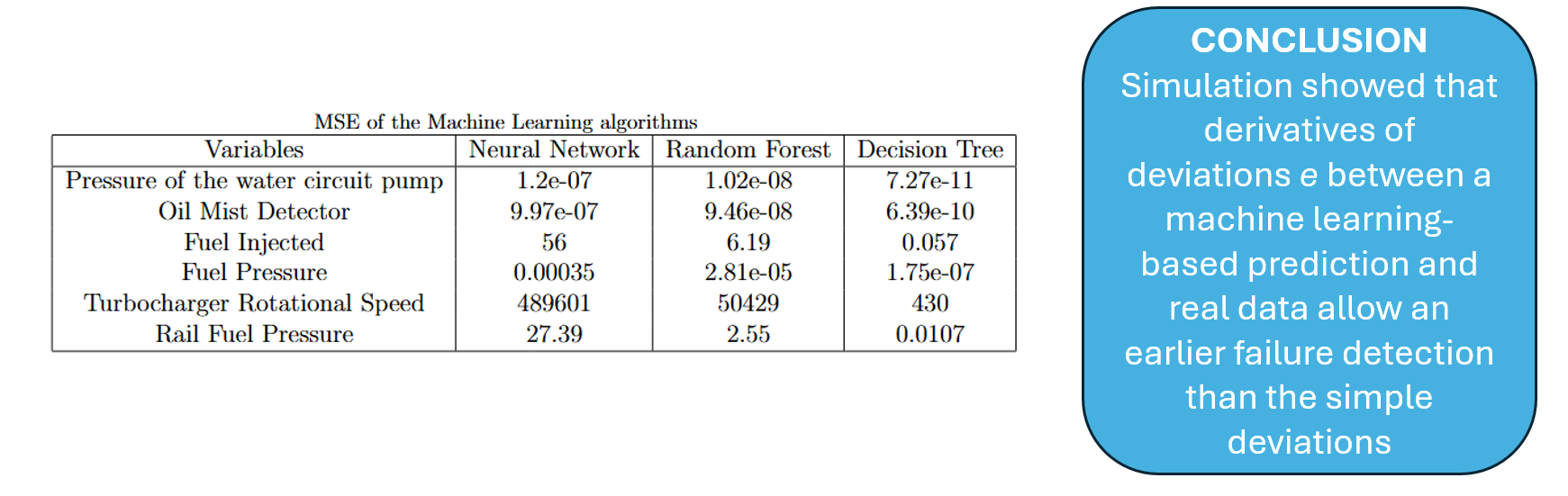}
\end{graphicalabstract}

\begin{highlights}
\item A 
Machine Learning-based methodology for early detection of catastrophic failures
\item {A novel detection strategy based on derivatives of deviation between predicted and expected behaviors}
\item Addressing the problem of lack of data by augmentation through a Variational Autoencoder 
\item Test of the proposed methodology by real data from a marine Diesel engine. 
\end{highlights}


\end{frontmatter}




\section{BACKGROUND AND MOTIVATION} \label{introduction}

{

Preserving or restoring machine operability is the main objective of maintenance 
\cite{why_and_when}. 
In the IFAC terminology, diagnosis is a crucial point of maintenance operations. It 
involves identifying the type, size, location, and timing of failures \cite{IFAC_terminologia}. 
This becomes particularly challenging for complex systems such as marine engines. 
For instance, in \cite{IFAC_failure}, a fault isolation and diagnosis framework was developed using cyber-agents, each dedicated to monitoring a specific subsystem of the engine 
and represented by algebraic equations. 
However good results {in fault diagnosis}, a pure model-based approach can lead to excessive 
simplifications and inaccuracies when applied to complex systems. Consequently, hybrid approaches have emerged, combining thermodynamic modeling with data-driven techniques such as neural networks \cite{paper_greco_2}.

In recent years, considerable attention has shifted from diagnosis to Prognostics and Health Management (PHM). Prognostics aims to estimate the Remaining Useful Life (RUL) of components, allowing timely {procurement} of spare parts and maintenance planning based on the current health status of the monitored engines \cite{LSTM_NTNU}. A common approach is the residual analysis, where deviations between current and expected healthy system behavior are monitored. An example is given by \cite{Puig}, in which residuals of a brushless DC motor were analyzed to predict failure time. Similarly, a Deep Learning (DL) algorithm based on a Long Short-Term Memory (LSTM) network was applied in a “learning from the field” approach \cite{Lecco_noi}. The aim was to forecast sensor signals and detect faults in advance, and address the 
challenge of shortage of historical failure data in the maritime field. 
\subsection{Data-Driven Condition-Based Maintenance}
Machine Learning (ML) and DL techniques have shown significant potential in PHM. Their success is due to the fact that data-driven approaches do not require a detailed physical representation of the monitored system, as model-based methods \cite{NTNU_2}. This is particularly advantageous in the case of marine Diesel engines, where physics-based models are often highly complex or unavailable because of trade secrets. However, some limitations remain because 
the large sets of ``labeled'' data {-- namely, distinct data associated to faulty and healthy conditions, respectively --} that are necessary to train supervised algorithms are often difficult to obtain \cite{acc_NTNU}. In many practical cases, the absence of faulty data restricts training of ML and DL {algorithms} only to healthy conditions, with anomalies detected as deviations from the learned normal behavior. In \cite{Ceglie}, an Artificial Neural Network (ANN) was trained on input variables unaffected by faults to predict the healthy behavior of a marine Diesel engine. { This work is based on a crucial issue, namely the identification of fault indicators.} 
{This concept, has been mentioned in } \cite{Rubio}, {where} a 
Diesel engine model was developed to 
simulate the most common failure scenarios and correlate sensor measurements with specific failure modes. {Despite the identification of key parameters and diagnosis of marine engine faults, the main drawback is the use of simulated data. Consequently, the simulated faulty behavior is close to the real one but never equal, as consequence False Positive or False Negative can be flagged. } 

While these strategies have proven effective for gradual degradation, far fewer studies have focused on catastrophic failures. They are 
sudden events that 
destroy the component, which will not continue to work \cite{Power_Grid}. These failures are often unpredictable and may have dramatic consequences for maritime safety. A tragic example is the collapse of the Baltimore bridge following the collision of a vessel that had suffered a power loss (see https://edition.\\cnn.com/2024/05/14/us/baltimore-bridge-collapse-ntsb-report/index.html). \\ 
In fact, catastrophic failures of propulsion systems result in loss of maneuverability. If they occur in rough sea or in proximity to obstacles, the risk of a deadly accident is high. 
This motivates research for the early detection of catastrophic events. In \cite{bearing_1}, Fault Tree Analysis was used to identify causes of bearing failures, while \cite{bearing_2} employed the Systems Theoretic Accident Model and Processes (STAMP) framework to analyze accidents and uncover underlying causes and human errors. More recent approaches leverage data-driven techniques: in \cite{bearing_3}, a Gated Recurrent Unit (GRU) network was used for real-time detection of outliers, emphasizing the algorithm’s response time; in \cite{bearing_4}, an Adaptive Estimator combined a Kalman Filter with residual monitoring to address the poor performance of standard Kalman Filters in abrupt events.

Despite the promising results of these methods, our study adopts a fully data-driven regression approach. In fact, model-based approaches for marine engines can suffer of significant approximation errors between predicted and real system dynamics {\cite{LSTM_NTNU,Lecco_noi}}. On the contrary, fully data-driven regression methods enable direct prediction of sensor measurements and comparison with observed values. Moreover, while GRU-based classification approaches 
provide binary labels (healthy/faulty), they lack insight into the dynamic transition between states. By focusing on regression, we aim to capture these transitions and provide a more informative and robust framework for the early detection of catastrophic failures.
Finally, we address the problem of the lack of data highlighted by \cite{acc_NTNU} through a DL-based data augmentation technique. 
}

\subsection{Research targets}
The final aim of early detection of failures in marine engines 
is to give information 
on time, 
such that 
the operators can shut down the engine in advance and protect it from huge damages. {In fact, the detected time of the failure does not correspond to the shut down of the engine. Assuming the presence of human operators on board, their intervention is not instantaneous once an alarm or warning is given. 
Since that, the action is performed too late, and the engine cannot be protected from total disruption, with the risk of unexpected loss of power and loss of maneuverability. Avoiding these situations is the core part of the current work. To this aim, the developed methodology aims to advise the onboard operator earlier than common warning and alarm logics that are already implemented. Even gaining a few seconds implies avoiding unexpected and dangerous situations.}

Firstly, we train three different ML algorithms to evaluate the {deviation} between the expected nominal behavior 
and the actual one
shown by sensor measurements. {The considered ML algorithms are an ANN, a Decision Tree (DT), and a Random Forest (RF) {because they are the most reliable and are widely applied in regression problems}. Then, the one showing the best prediction of expected behavior is selected. Note that the performance has been evaluated by the Mean Squared Error (MSE).}
Secondly, we evaluate the first and 
second derivative 
of the {deviation mentioned above}. 
The second derivative 
was also used in \cite{acc_NTNU} calculated 
to detect whether 
engine 
failures occur together. 
More in detail, in \cite{acc_NTNU}, the authors proposed an unsupervised learning to apply different reconstruction algorithms 
and evaluate the reconstruction error, such that the estimation of the time-step showing 
the highest acceleration in 
the anomaly score identifies 
the fault. 
On the contrary, we propose a supervised learning by a ML algorithm whose output provides the expected healthy engine behavior. 

{Although the unsupervised approach is recommended in the absence of labeled data, as in our case \cite{NTNU_2}, supervised learning allows a better localization of failures {\cite{Ceglie}}. In fact, the detection by an unsupervised algorithm usually works by using sensor measurements as inputs and their reconstruction as outputs (see \cite{acc_NTNU}). Consequently, the fault is detected by computing an aggregate index providing the deviation between the reconstructed variables and the original inputs, for example, the Mean Squared Error. Then, the fault detection is related to a global unhealthy behavior of the engine. 
}

On the contrary, the inputs of supervised learning algorithms are often data that are not affected by faults, which means that the outputs are always the sensor measurements expected for a normal (healthy) behavior. Then, observing the actual sensor measurements, which may differ from the expected behavior, allows us to easily identify the faulty component.

We developed the current study with the assumption that, even if the {deviation} 
is below the 
threshold, which is usually defined for warning or alarms (5\%--10\% of the expected value), the derivatives 
are more suitable to 
early detect 
sudden failures, especially when they show a rapid and constant increase. 
{Furthermore, in \cite{acc_NTNU} and {\cite{ECC_noi}}, the authors prefer the second derivative of the {deviation} with respect to the first} because the latency of the physical components makes the second derivative a better indicator. 
{In this work, we propose to check both the derivatives simultaneously, as they give a more meaningful {and complete} indication.} {Namely, the derivatives provide information on how fast the deviations are changing, giving us the opportunity to visualize if something abrupt is happening in the engines. Since the detection time of catastrophic failures is critical and even few seconds in advance can be very helpful, 
the derivatives of the deviations are chosen.}


{Finally, the current work gives its contributions even to other problems affecting the maritime sectors: the lack of data and the noise reduction. To this aim, we apply a Variational AutoEncoder (VAE) for data augmentation, thereby increasing the amount of data to train the ML algorithms. Then, we use the average mean on a sliding time window to reduce the noise effect. 
}
{To make more realistic simulation tests, the proposed method has been tested by taking advantage of real data, provided by an Italian company which supports the current study, namely Isotta Fraschini Motori S.p.A., a Fincantieri company (see https://www.isottafraschini.it/en/). 
Due to a confidentiality agreement, the data will be shown in a normalized way. 
The acquired catastrophic failure is a bearing collapse.}

We finally remark that the contributions of this study 
are the following: 
\begin{itemize}
\item An effective methodology to detect catastrophic failures in a marine Diesel engine; to the best of 
our knowledge, no similar technique 
was proposed for catastrophic failures.
\item {The data augmentation technique based on the { VAE algorithm}}; to increase the amount of data, hence to prevent the problem of lack of data, which is typical in catastrophic failures of marine engines.
\item The possibility of applying the strategy 
on { real sensor data from a catastrophic failure, such as a bearing collapse.} 
\end{itemize}

The 
rest of the paper is divided as follows. Section \ref{back} introduces the required preliminaries and background; 
Section \ref{methodology} presents the 
methodology; Section \ref{data} describes how data are obtained from the monitored asset; Section \ref{simulation} shows the simulation results; 
finally, Section \ref{conclusions} draws the conclusions.

\section{PRELIMINARIES} \label{back}
\subsection{Catastrophic Failures}
{Ageing failures are the most common kind of anomaly behaviors that can be observed during the working life of an engine. As the name suggests, they are due to the passage of time, namely to the wear. As time-dependent behaviors they can be simply modeled as follows: 
\begin{equation} \label{time}
    y_{f}(t)=y_h(t)+k(t) \, .
\end{equation}
This equation represents a so-called additive failure: the time-dependent fault coefficient $k(t)$ is added to the healthy value $y_h(t)$ to obtain the faulty value $y_f(t)$. Other kinds of failure can be modeled as ``multiplicative" faults:
\begin{equation} \label{time_2}
    y_{f}(t)=y_h(t) \cdot k(t) \, .
\end{equation}
Independent of the way in which an ageing failure is modeled, its behavior is the same: it shows a slow and constantly {increasing} deviation of $y_{f}(t)$ from the expected value until it reaches the maximum allowable value.

}
Before reaching this maximum, 
the overall 
system may continue to operate at lower performance until the maintenance action is necessary. For example, a degraded faulty marine engine can still work but provides a 3\% increase of emissions, {at least} \cite{emissioni}. 

On the contrary, a machinery suffering a catastrophic failure immediately stops 
and 
needs 
urgent and expensive reactive maintenance or 
total replacement. 
{Catastrophic failures can easily be modeled as follows}:

\begin{equation}
k(t) =
\begin{cases} 
    0 & \text{if } t<t_{f}, \\
    F & \text{if } t \geq t_{f}.
\end{cases}
\end{equation}
where $t_{f}$ is the failure occurrence time and $F$ is the failure value. Note that the catastrophic failure is assumed to provide a constant value of $k(t)$ to remark 
that it is 
irreversible. 
{ In this case, the failure coefficient $k(t)$ changes suddenly from 0 to the faulty value, meaning the total disruption of the asset.}
After its occurrence, the engine loses all the 
ability to function, unlike with other faults when faulty operating conditions are still possible.

\begin{figure}[H]
      \centering
      \includegraphics[width=\linewidth]{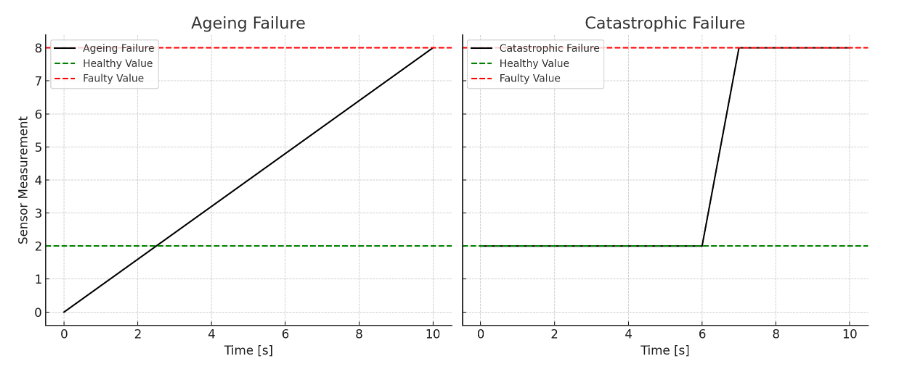}
      \caption{Difference between aging and catastrophic failures}
      \label{fig.failure}
   \end{figure}

A simple representation of the main differences between the two types of failure is given by Fig.~\ref{fig.failure}. 
As shown, the variable given by a sensor measurement slowly increases from a healthy to a faulty state in the case of an ageing failure. Instead, the variable rapidly, or even abruptly, changes in the case of a catastrophic failure. This rapid variation can not be detected by traditional methods, whereas the method we propose is conceived for fast changes.

The main difficulty 
in detection of a catastrophic failure is the velocity with which it happens. 
{ For this reason, the only way to act is to \textit{early} detect the catastrophic failures, namely to generate alarms or warning signals even a few seconds before, thus anticipating the intervention of the operators and giving them an adequate time to shut down the engine and avoid sudden and unexpected loss of maneuverability. }

\subsection{First ML algorithm: the Decision Tree}

A Decision Tree (DT) is a supervised ML model applied to make decisions based on a 
collected dataset. 
It can be trained both for classification, that is, to give labels to the samples, or for regression to predict the expected values.
{Herein}, the DT is trained to  
predict the expected sensor measurements, namely the 
measurements not affected by faults. {In machine learning, the DTs are more applied for classification than for regression problems. However, the DT is suggested in applications where a high robustness to noise is required \cite{ECC_noi,Algeria}. In fact, even if the generic sensor measurement can be defined as in (\ref{time}), the true real values $y_r(t)$ read by sensors are affected by noise $n(t)$: 

\begin{equation}
    y_m(t)=y_r(t)+n(t)
\end{equation}



The DT is composed by nodes {and arcs}: 
the input nodes are called roots and represent the first decision taken based on input data; the internal 
decision nodes 
determine the flow of the decision process; 
the final nodes are called leaves and represent the 
predicted values; an arc between two nodes represents the result of each decision. 
One of the most important properties of a DT is the entropy, 
i.e. the level of uncertainty in assessing a variable in a decision process: 

\begin{equation} \label{entropy}
    E(D)=- \sum_{x \in X}{p(x) \, log_{2}p(x)} 
\end{equation}
with $E(D) \in [0,1]$, where $D$ is the dataset, $X$ a class in the dataset $D$, $x$ a variable, and $p(x)$ is the probability that $x \in X$. The entropy 
measures the “disorder” related to a split: the lower is the entropy after a 
decision, the more effective the training is. 
{Fig.~\ref{fig.DT} gives an overview of the DT structure for regression problems. In the figure, $R$ and $P$ are the rotational speed and power, respectively, i.e. the inputs of the DT; $R1, R2, R3,...,RN$ and $P1, P2, P3,..., PN$ are the rotational speeds and powers which the DT chooses during the training phase, as references for its decision process. Depending on the values of the inputs and on the decision rules that the DT learns, the decision process flows through different steps (displayed in green color) until the prediction is made.
}
      \begin{figure}[H]
      \centering
      \includegraphics[width=\linewidth]{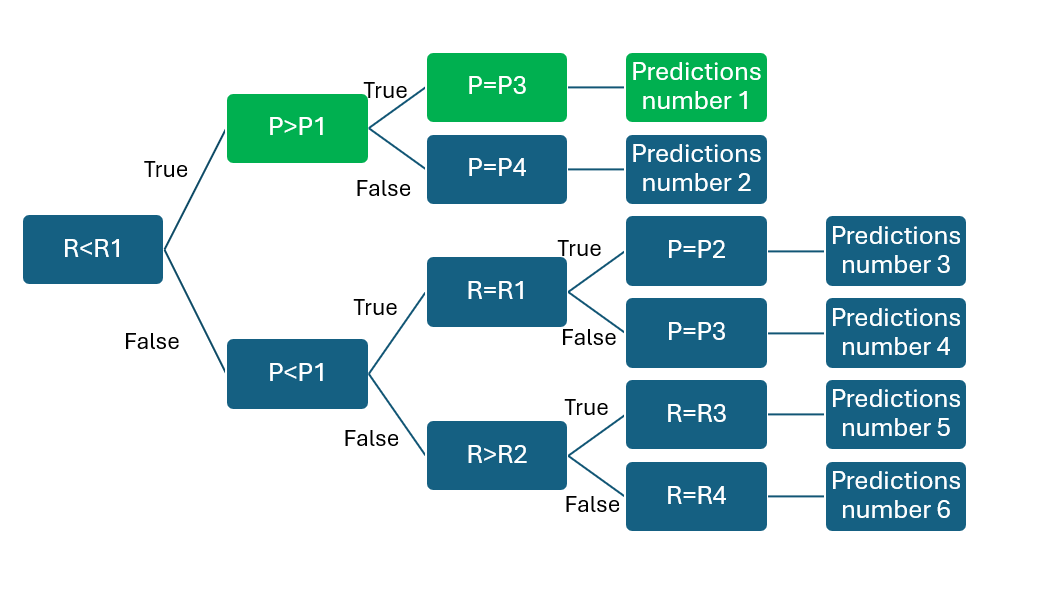}
      \caption{Decision tree {for the regression problem}}
      \label{fig.DT}
   \end{figure}

\subsection{{Second ML algorithm: the} Random Forest}
A Random Forest (RF) is an ensemble supervised algorithm composed by 
a combination of DTs \cite{Random_Forest}. 
%
%
The ensemble of DTs 
improves the performance of a single DT by aggregating several predictions to find the most widespread result. 
The first step of its realization is the method called bagging, or bootstrap aggregation 
of datasets, which is obtained from an initial dataset by a random sampling with replacement.  
Basically, elements are taken from the original dataset and used to train one of the Decision Trees in the Forest. Then, these elements are placed back in the original dataset, such that they can be re-used for other trees.
After that, during the learning phase, another bagging is performed to 
find the specific features that allow the node split and a reliable prediction to obtain the correct output.
Averaging the tree outputs or selecting the output from the highest number of trees provides the final output.

The decision process of the single decision tree is based on the \textit{Gini} index:

\begin{equation}\label{eq7}
G(t) = 1 - \sum_{i=1}^{N} p_i^2 
\end{equation}
which evaluates the probability that a sample belongs to a class, where $N$ is the total number of classes and $p_i$ is the fraction of elements belonging to class $i$ in a specific node. 

Since the RF is a combination of DTs, the entropy is a key index even here, and it is computed as in (\ref{entropy}). 
 

It is remarked that in case of a regression problem, 
the output is evaluated by 
\begin{equation}\label{eq9}
\hat{y} = \frac{1}{T} \sum_{t=1}^{T} \hat{y}_t
\end{equation}
where $\hat{y}_t$ is the output of a DT in the Forest and $T$ is the total number of trees.

\subsection{{Third ML algorithm: the} Artificial Neural Network}
A traditional Artificial Neural Network (ANN) 
consists of different interconnected hidden layers, where neurons process the data to produce an output, as shown by Fig.~\ref{fig:NN_structure}. 

\begin{figure}[H]
  \centering
  \includegraphics[width=\linewidth]{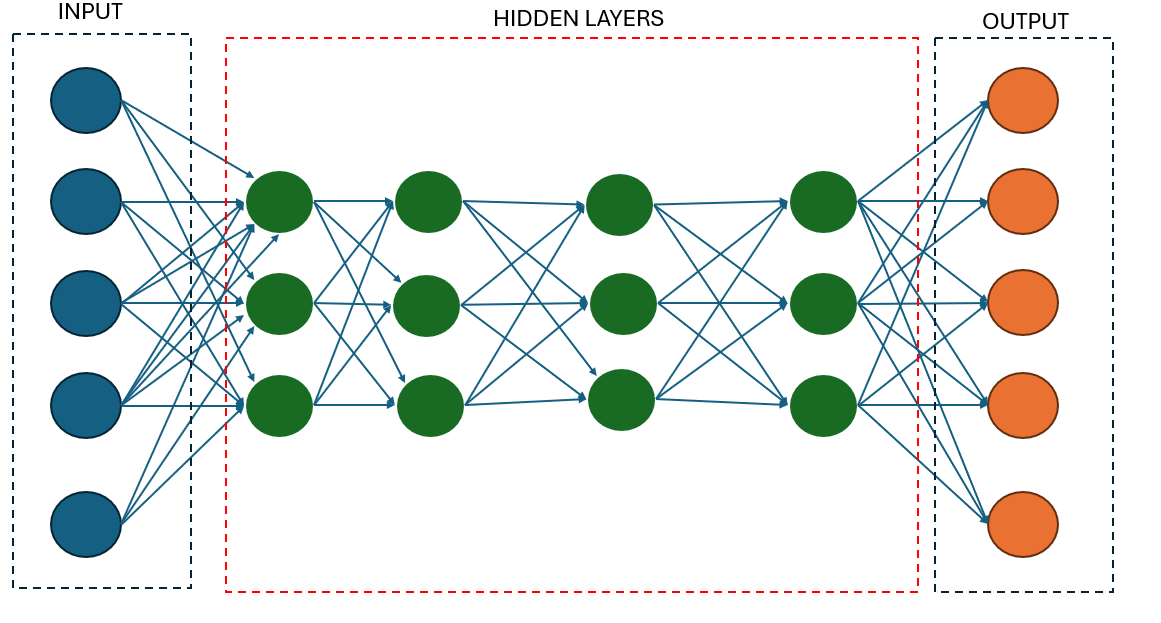} 
  \caption{Structure of Artificial Neural Networks}
  \label{fig:NN_structure}
\end{figure}

{
The input values are multiplied by the weights in the hidden layers. Initially, the weights are set randomly. The multiplication results are combined together to produce the neuron output. Finally, once the information arrives to the final layer, the ANN provides its outputs. 
This process is called forward propagation. However, in the training phase, there is also a 
backward propagation, 
during which the weights are optimized. 
More details are in \cite{Reti_Neurali}. 
The complete ANN algorithm is described by the following equations:

\begin{equation} \label{eq1NN}
y = f\left( \sum_{i=1}^{n} w_i x_i + b \right)
\end{equation}


\begin{equation} \label{eq3NN}
h_j = f\left( \sum_{i=1}^{n} w_{ij} x_i + b_j \right)
\end{equation}

\begin{equation}\label{eq4NN}
y_k = f\left( \sum_{j=1}^{m} w_{jk} h_j + b_k \right)
\end{equation}

\begin{equation}\label{eq5NN}
\Delta w_{jk}= \lambda \varepsilon_k h_j
\end{equation}
where {(\ref{eq1NN}) computes} the output of the neural unit, $x_i$ are the inputs, $w_i$ are the weights, $b$ indicates the bias and $f$ the activation function. In (\ref{eq3NN}), $w_{ij}$ is the weight associated to the $i$-th neuron input $x_i$ and the $j$-th neuron output $h_j$; $b_j$ is the bias of the $j$-th neuron. In (\ref{eq4NN}), $b_k$ is the bias of the $k$-th output; $w_{jk}$ is the weight associated with the $j$-th hidden layer input and the $k$-th output. 
Eq. (\ref{eq3NN}) and (\ref{eq4NN}) represent the feedforward operations for the hidden layer and the output layer, respectively. 
Here, $h_j$ is the neuron output of the $j$-th hidden layer; $y_k$ is the $k$-th output. Eq. (\ref{eq5NN}) describes the backpropagation: $\lambda$ is the learning rate and $\varepsilon_k$ the error between the network output and real data. It is remarked that the backpropagation consists \textcolor{red}{(of?)} on updating the weights to increase the prediction capability of the ANN. 

It is worth noting that, unlike the RF and DT, the data managed by the ANN must be standardized as follows: 

\begin{equation}
    x_Z=(x-\mu_\textbf{x})/\sigma_\textbf{x}
\end{equation}
where $x_Z$ is the standardized value, $x$ the acquired single value, $\mu_\textbf{x}$ and $\sigma_\textbf{x}$ the mean and the standard deviation of the acquired variables. It is remarked that the mean and the standard deviation are evaluated by using the training dataset only. }

\subsection{{Data Augmentation by the} Variational Autoencoder}
{ The Variational AutoEncoder (VAE) is an 
algorithm which is a variant of the traditional AutoEncoder (AE), which was proposed for the first time by \cite{VAE_primi}. In fact, while a traditional AE is mostly able to reconstruct the input variables thanks to the encoder-decoder structures, the VAE learns the {probabilistic} distribution of the input features by extracting the latent representation of the input data. In this way, the VAE 
is not only able to reconstruct the input, as a traditional autoencoder, but also to give new ``artificially" generated data similar to the input ones as output.

If $z$ is the latent space obtained after the encoder, it can be represented by the following equation: 
\begin{equation}
    z=f_e(z|x)
\end{equation}
where $f_e$ is the encoder function and $x$ are the input data. 
Then, the decoder reconstructs the input through the formula: 
\begin{equation}
    g=f_d(x|z)
\end{equation}

Moreover, the training of the VAE has as objective the minimization of the loss function:

\begin{equation}
    L=(-1)*D_{KL}(f_e(z|x)||p(z))+E_{f_e(z|x)}[log(f_d(z|x))]
\end{equation}
where $D_{KL}$ is the Kullback-Leibler divergence between the probabilistic distribution $f_e(z|x)$ and its approximation $p(z)$, which is used as a regularization factor, and $E_{f_e(z|x)}[log(f_d(z|x))]$ is the log-reconstruction likelihood{, namely the distance loss between the two distributions of the Kullback-Leibler divergence}. 
The $z$ variable is obtained through a reparameterization: 
\begin{equation}
    z=\mu+\sigma \varepsilon
\end{equation}
where $\epsilon$ is normally distributed, {$\mu$ and $\sigma$ are the mean and standard deviation of the real measurement values}.

Finally, the VAE structure is composed of several interconnected layers, as a traditional ANN. Fig.~\ref{fig:VAE} gives a simple overview of the VAE structure. 
Similarly to the ANN, even the VAE inputs need to be standardized. 

\begin{figure}[H]
  \centering
  \includegraphics[width=\linewidth]{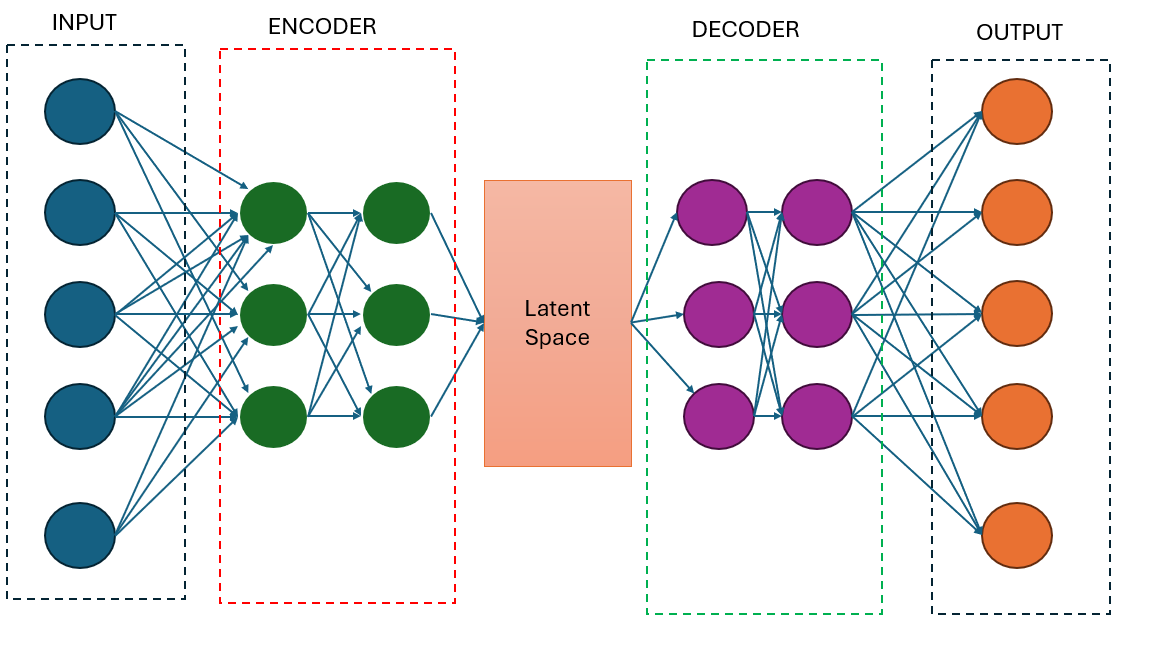} 
  \caption{Structure of the VAE}
  \label{fig:VAE}
\end{figure}

}

\section{THE PROPOSED METHOD } \label{methodology}
{This section gives an overview of the method proposed to early detect catastrophic failures. Each subsection describes parts and details of the followed procedure. 
For a better understanding, the results of each step in the procedure will be shown and explained with reference to the specific catastrophic failure of bearings in the considered engine. Details on the specific failure will be given in the following section.}
\subsection{Pre-processing and Variables Selection}
{

A pre-processing phase is necessary before using the data to train the chosen ML algorithm. In fact, the sensor measurements often contain outliers, non-realistic data, especially data acquired in non-nominal conditions, such as low power. Then, an accurate pre-processing is required to maintain only realistic data and train the predictive algorithm correctly. With this goal, all data that are not useful, such as those with negative or infinite values, are 
removed. 

The next step is to properly choose the variables, namely the sensor measurements, that are suitable to early detect the considered catastrophic failure, namely the bearing collapse. The aim is to choose the most selective variables among the available acquired ones. Such variables should be the most appropriate to detect the failure. For that reason, the percentage deviation between the healthy and faulty values is computed for each sensor measurement. Then, the variables with the highest deviations are selected for detection. 
An overview of the {deviations displayed by all the considered variables} is given in Fig.~\ref{fig.Bar_Graph} for the analyzed case. 

         \begin{figure}[H]
      \centering
      \includegraphics[width=\linewidth]{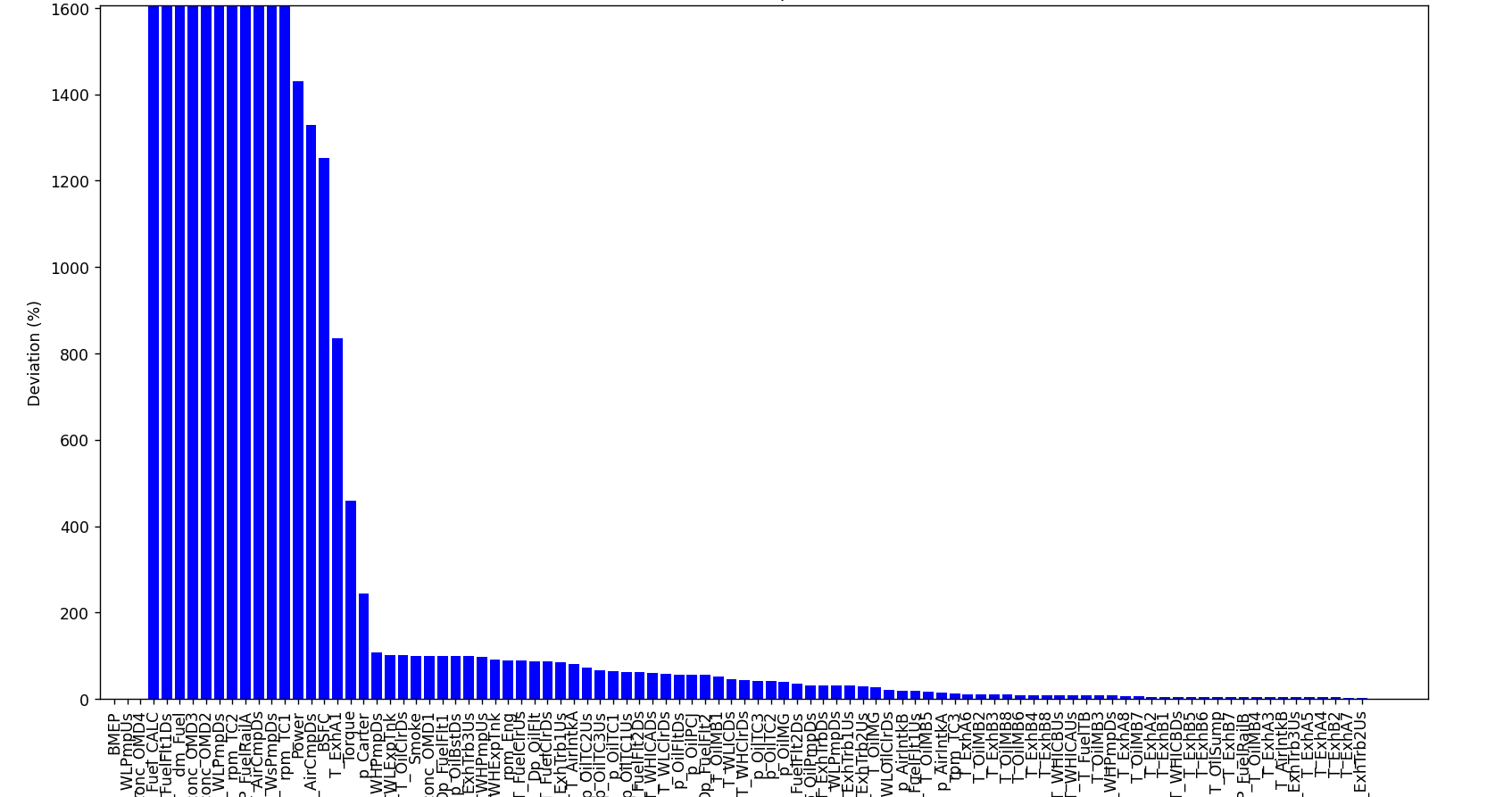}
      \caption{Deviations of variables from expected values during the catastrophic failure 
      }
      \label{fig.Bar_Graph}
   \end{figure}

Then, the variables chosen to detect the bearing collapse are the following: 
\begin{enumerate}
    \item[a.] The pressure generated by the pump in the water circuit; 
    \item[b.] The fuel pressure in the injection circuit;
    \item[c.] The fuel pressure coming out of the nozzle (injector output), measured at the injection time;
    \item[d.] The oil mist measured by three sensors;
    \item[e.] The common-rail pressure;
    \item[f.] The rotational speed of turbo-chargers. 
\end{enumerate}

The enumerated variables will be used as outputs of the ML algorithms to be compared. Namely, each algorithm will be trained to predict the expected (healthy) values based on the engine working conditions (rotational speed and power). 
}

\subsection{Noise Reduction}
{The data acquired from the real asset are highly affected by noise, due to both internal noise from sensors and environmental noise. For that reason, noise reduction is necessary even to reduce the peaks that can occur during the computation of the derivatives of the deviations from healthy values. 
We decided to apply a simple moving average with a sliding window of 300 seconds because of ease of implementation and low computational cost. The sliding window 
has been evaluated {by trial and error} based on the acquired dataset. 
}

\subsection{Data Augmentation}
{ The lack of available data from new-generation engines or engines that are protected by a trade secret is one of the main problems in the maritime sector \cite{IFAC_failure, acc_NTNU}. 
To address this problem, we proposed the application of the DL algorithm represented by the VAE to ``artificially" augment the data necessary to train the ML algorithms. Data augmentation is a widely applied technique, which is also used by the authors of \cite{LSTM_NTNU,LLM}. The purpose of data augmentation is to increase the generalization capacity of the predictive algorithm. In fact, the lack of publicly available data, the restrictions imposed by trade secrets, and the classified information from military applications make it difficult to find sufficient data, both in healthy and faulty conditions. This occurs in all data-driven learning in which ML and DL algorithms are applied. For that reason, the VAE has been introduced to increase the healthy data used for training. More in details, 37109 data were acquired during the healthy load profile simulation. The VAE allowed us to obtain the same amount of data, such that 74218 values were used to train the algorithms. Note that both the inputs (engine rotational speed and power) and the outputs (the variables identified in the previous section) have been augmented. 
}

\subsection{Selection of the Machine Learning Algorithm} 
{
We applied three of the most popular ML algorithms used for fault detection: the ANN, the DT, and the RF. The ANN has been chosen as it is widely applied in this field, because of its capacity to recognize non-linear correlations between inputs and outputs. Its main limit is the sensitivity to the noise. To address this issue, DTs are often applied as more robust. For that reason, a comparison between an ANN and a DT is a good starting point for the choice of the best predictive algorithm. Moreover, the RF has been considered because it is an ensemble of DTs, then it can amplify the advantages of a single DT, such as a better generalization capacity. Once the algorithms are trained on the augmented data, the one with the lowest MSE is chosen for the early detection of catastrophic failures. 

}
\subsection{Detection of Catastrophic Failures}
The main reason for using the 
derivatives {of deviations from healthy behaviors} 
is that catastrophic failures occur in few seconds, such as the time between the hearing of an abnormal sound and the seeing of smoke coming out of the engine \cite{bearing_2}. Then, the key idea 
is to provide information as soon as possible {and a simple deviation from the healthy behavior may not detect an abrupt change far in advance}. Instead, the (first and second) derivatives 
are more suitable than the simple deviation 
if something is happening in a few seconds, because they provide information about how fast the {deviation and its first derivative} are increasing or decreasing (speed and acceleration of change). 

The proposed method 
uses only healthy data of a marine Diesel engine to train the ML algorithms. The algorithms' inputs are the rotational speed (rpm) and power (kW) for each 
operating condition, for which each trained algorithm generates the expected sensor measurements. 
Once the different algorithms are trained, the one with the highest performance is selected and applied to the engine. 
Called $N$ the number of healthy acquisitions of the $n$ sensors installed on the 
engine, the whole dataset is divided into two parts. The first part (75\% of data) is used for training, 
the second part (25\% of data) for testing the algorithms. 
%
The MSE 
is used to evaluate the prediction performance. 

Once the prediction algorithm is chosen, its
outputs $\hat{y}_{i}$ are compared with the 
measurements $y_{i}$ to evaluate the {deviations} $e_i(t)$: 

\begin{equation} \label{eqerr}
e_{i}(t)= \frac{y_{i}(t)-\hat{y}_{i}(t)}{\hat{y}_{i}(t)} \quad \text{for } i=1,2,...,n.
\end{equation}



Note that, since the predictive algorithm 
is trained on healthy values only, 
its outputs are expected to be the sensor measurements for healthy conditions. 
Then, the first and second derivatives 
of the {deviation} are evaluated: 
\begin{equation} \label{eqvel}
v_{i}(t) = \frac{e_{i}(t+1)-e_{i}(t)}{\Delta T} \quad  \text{for } i=1,2,...,n
\end{equation}
\begin{equation} \label{eqacc}
a_{i}(t) = \frac{v_{i}(t+1)-v_{i}(t)}{\Delta T} \quad \text{for } i=1,2,...,n.
\end{equation}
where $\Delta T$ is the {data acquisition} time-step.

{A crucial step for the correct and early detection of catastrophic failures is the setting of suitable thresholds. In fact, the thresholds established for the deviations are given by years of experience or can be derived from the available technical and scientific literature. Moreover, since no ML algorithm, however well trained, is able to perfectly predict the real behavior, a deviation is always present. {On the contrary, no literature or previous knowledge exist for thresholds related to the first and second derivatives of the deviations. For that reason, we propose the following solution: the first and second derivatives of the deviations are evaluated while predicting the trend of a healthy engine load profile. Then, the maximum values of the derivatives evaluated on this profile are chosen as thresholds for the early detection of catastrophic failures. }

An overview of the 
method is given by the Algorithm \ref{Alg1} and by Fig.~\ref{fig:method}.

        
\begin{algorithm} \label{Alg1}
\SetAlgoLined
\KwResult{Training of the ML algorithm and evaluation of the derivatives 
of the {deviations} {to detect} catastrophic failures}
\vspace{5pt}

 \textbf{Initialization}:\\
 Input matrix $X$: power and rotational speed corresponding to $w$ sensor measurements;\\ 
 Output matrix $Y$: healthy $w$ sensor measurements from the engine for 50 variables. 

\textbf{Training}:\\
X and Y are divided into:\\
$X_{tr}$, $X_{te}$=$X.divide(0.75, 0.25)$

$Y_{tr}$, $Y_{te}$=$Y.divide(0.75, 0.25)$

Model.Training(input=$X_{tr}$, output=$Y_{tr}$); \\
Performance.Evaluation(output=$\hat{Y}_{tr}$, real.output=$Y_{tr}$, metrics=$MSE$); \\

\textbf{Test}: \\
Model.Prediction(input=$X_{te}$); \\
Performance.Evaluation(output=$\hat{Y}_{te}$, real.output=$Y_{te}$, metrics=$MSE$); \\

 \vspace{10pt}

\textbf{Threshold Definition}: \\
\vspace{5pt}
Model.Prediction(input=$X_{healthy}$); \\
$V_{healthy}(t)$ first derivative computation; \\
$A_{healthy}(t)$ second derivative computation; \\
\For{each column of $A(t)$ and $V(t)$}{
$v_{threshold}=max(V(t))$; \\
$a_{threshold}=max(A(t))$;

}

\textbf{Application of the algorithm 
to the engine
}:\\
 \vspace{5pt}
 \For{each discrete time $t = k \, \Delta T$}{

$E(t)$: matrix of the errors; \\
$V(t)$: matrix of the first derivatives of errors; \\
$A(t)$: matrix of the second derivatives of errors;\\

  \If{$V_{n_im_i}(t=t_n)$ or $A_{n_im_i}(t=t_0) > v_{threshold}$ or $a_{threshold}$}{
   switch off the engine 
   }
 }
 \caption{Training and test of the ML algorithm and computation of the derivatives of deviations}
\end{algorithm}


         \begin{figure}[H]
      \centering
      \includegraphics[width=\linewidth]{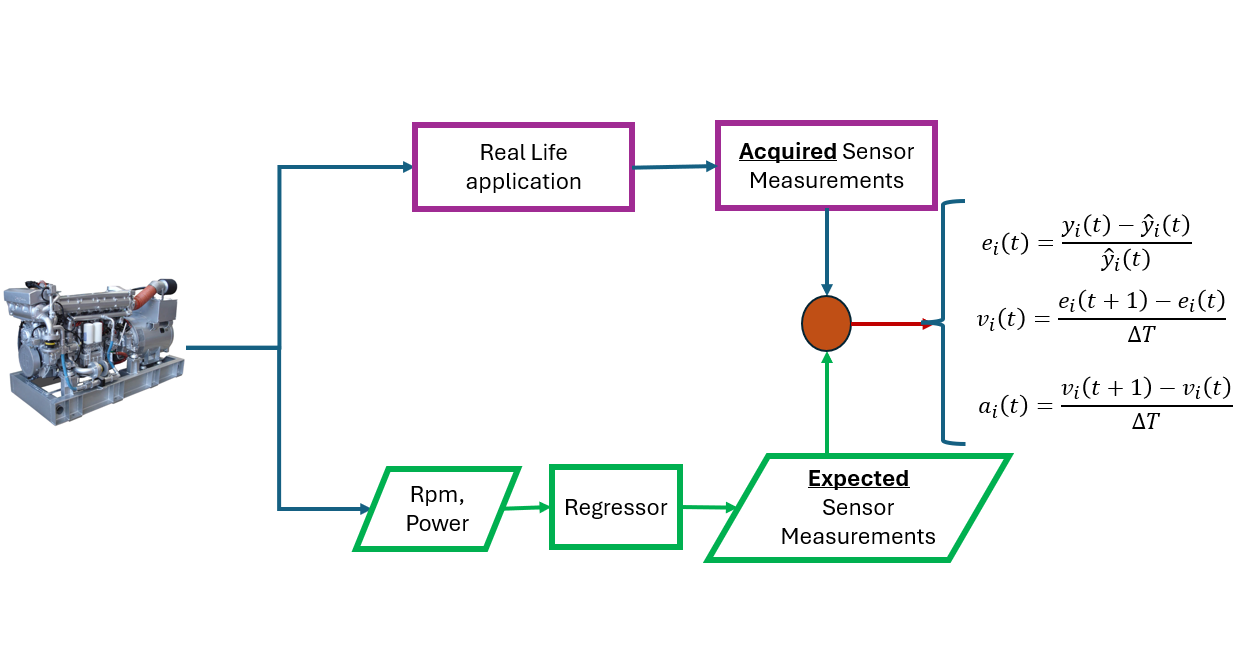}
      \caption{Overview of the proposed method. 
      }
      \label{fig:method}
   \end{figure}


{
It is important to remark that the proposed method is applicable for the early detection of all catastrophic failures, but, as in all the data-driven approaches, the results 
are strongly dependent on the application and on the observed catastrophic failures. More details on the current case study will be given in the following section.

}

\section{DATA FROM A FAILED MARINE ENGINE} \label{data}

{The strength and robustness of the method proposed for early detection of catastrophic failures also rely on validation using real data acquired from sensors deployed in the engine. 
The monitored asset is a marine Diesel engine composed of sixteen cylinders and three supercharging groups. Both the propulsive and generator set applications are suitable for the designed engine. However, 
due to a confidentiality agreement with the industrial partner involved in the current research (see https://www.isottafraschini.it/en/), no more details can be publicly provided. 

The monitored engine suffered a bearing collapse during simulation of a load profile: namely, the generated power was expected to change from a minimum to a maximum value and then back to the minimum, but the sudden catastrophic failure occurred once the maximum was reached. Then, the installed warning and alarm sensor logic gave an alarm and the operator shut down the engine. However, due to the late intervention, the engine was totally disrupted. 

An overview of healthy and run-to-failure 
behaviors is given in the Fig.~\ref{fig.Power}-\ref{fig.WL}. We remark that the data shown have been normalized to preserve the confidentiality of the asset.
The normalization has been applied by the following equation: 
\begin{equation}
    x_{aN}= \frac{x_a-min(x_a)}{max(x_a)-min(x_a)}
\end{equation}
where $x_{aN}$ is the normalized value, $x_a$ is the real value, $min(x_a)$ and $max(x_a)$ are the minimum and maximum values of the acquired {sensor measurements} of a variable, say $a$.
Note that, although only three variables are shown for lack of space, more than ninety sensors are currently installed on the asset.

      \begin{figure}[H]
      \centering
      \includegraphics[width=\linewidth]{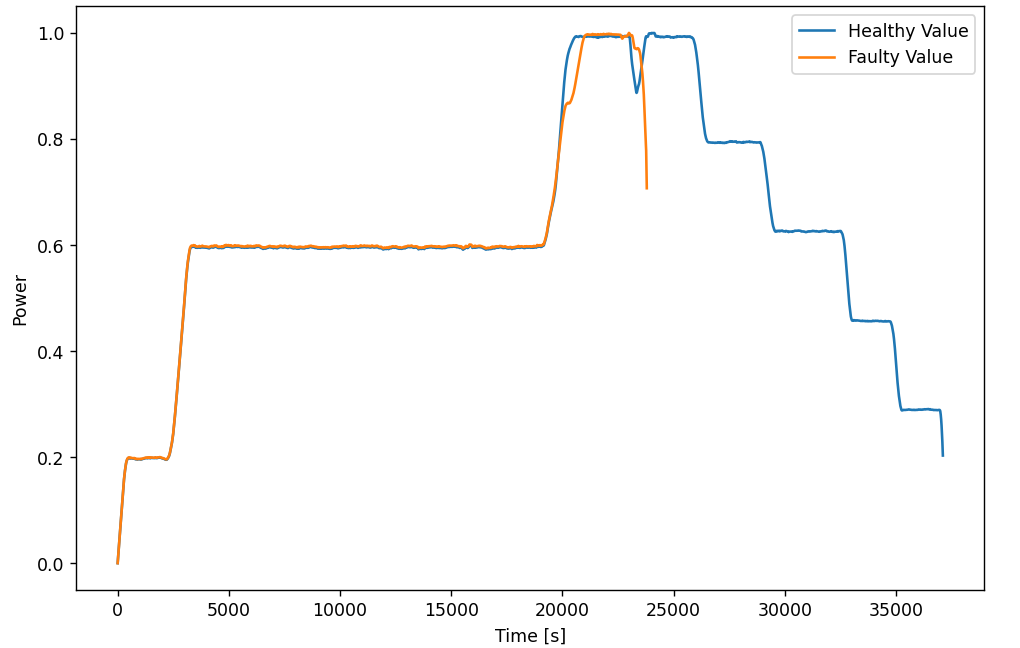}
      \caption{Engine power 
      }
      \label{fig.Power}
   \end{figure}

         \begin{figure}[H]
      \centering
      \includegraphics[width=\linewidth]{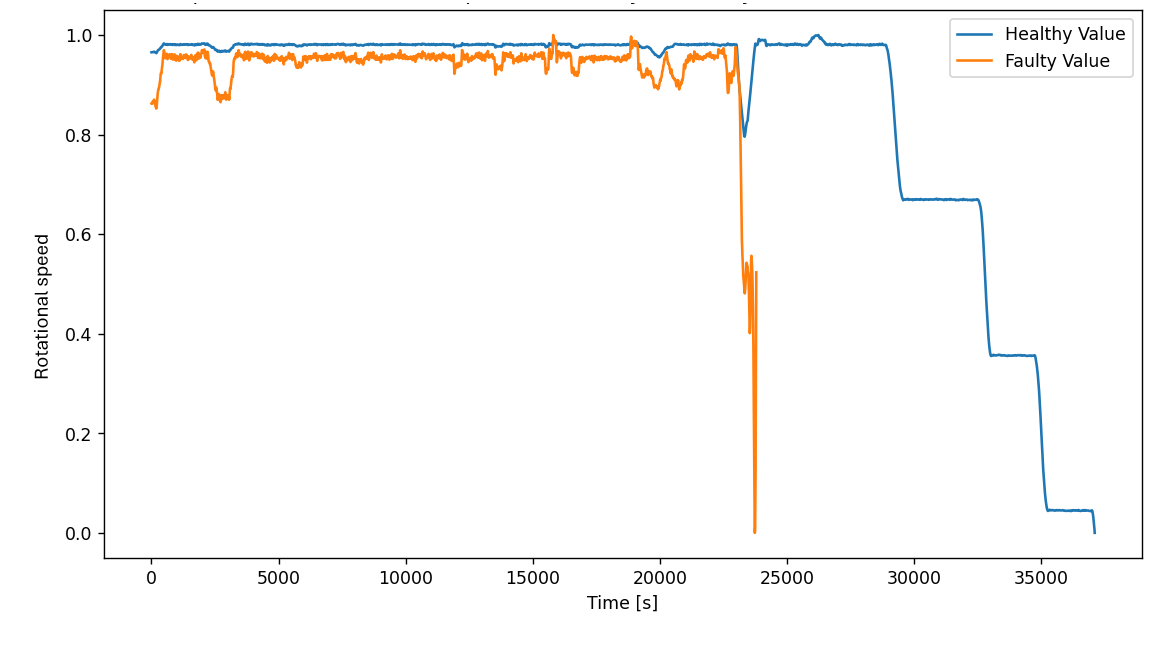}
      \caption{Engine rotational speed 
      }
      \label{fig.RPM}
   \end{figure}

         \begin{figure}[H]
      \centering
      \includegraphics[width=\linewidth]{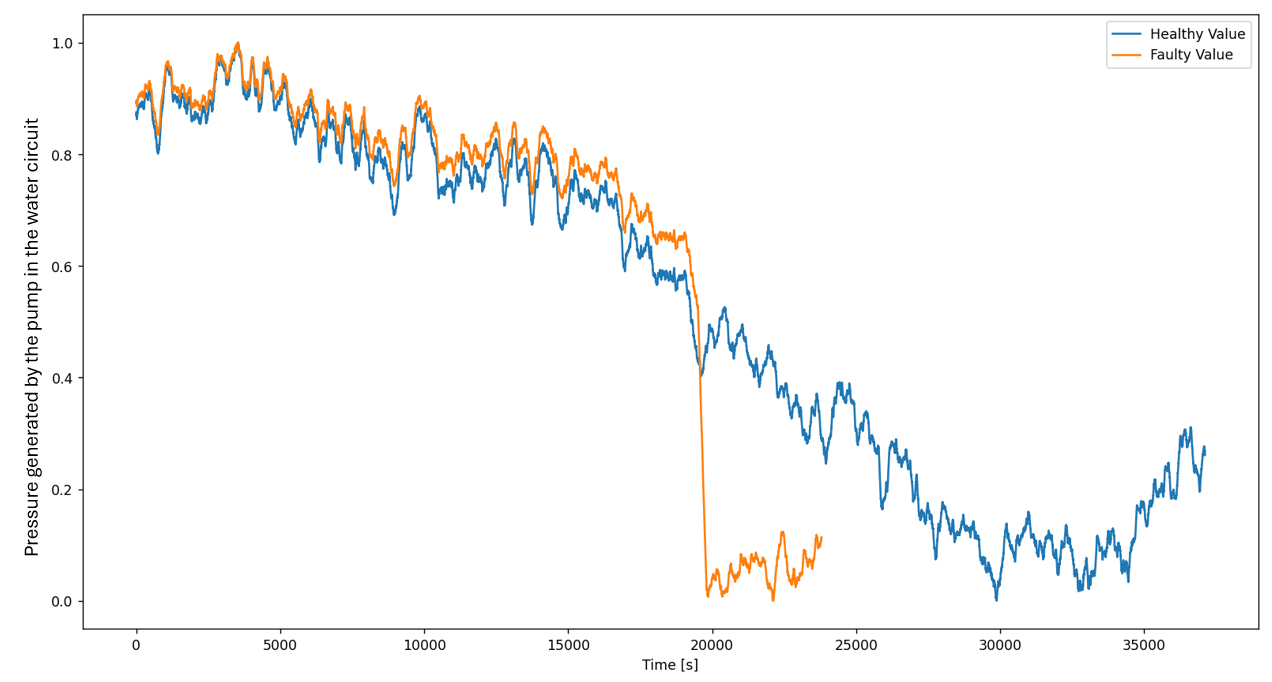}
      \caption{Pressure generated by the pump in the water circuit} 
      \label{fig.WL}
   \end{figure}

}

\section{TRAINING AND FAILURE DETECTION: RESULTS AND DISCUSSION} \label{simulation}
{In this section, we show the results in simulating the detection of a catastrophic failure by a predictive algorithm trained on this purpose. 
In this way, we focus exclusively on the core part of the proposed method. Simulation has been performed by Python programming language and with a computer with an Intel(R) Core(TM) i7-1065G7 CPU processor and 8.00 GB of RAM.}

{ 
The first results shown are those related to the data augmentation and training of the Machine Learning methods. 
The VAE hyper-parameters are shown in Table~\ref{tab:VAEhyp}. It is remarked that the hyper-parameters have been tuned by trial and error to reduce the VAE reconstruction loss, which was {2.792} during the training phase. 

\begin{table}[h]
  \centering
\caption{VAE hyper-parameters}
\label{tab:VAEhyp}
  \begin{tabular}{|c|c|}
    \hline
    Encoder activation function & \textit{ReLu}\\
    Encoder layers& 32\\
    Decoder activation function& linear \\
    Decoder layers & 32\\
    Training epochs & 400 \\
    \hline
  \end{tabular}
\end{table}

Then, the next step is the ML algorithms training. 
The total amount of data has been divided into a training set (75\%) and a test set (25\%). More in details, the training has been conducted to reduce the MSE. 
The training and test performance are shown in Tables~\ref{tab:Train}--\ref{tab:Test}. 

\begin{table}[h]
  \centering
\caption{Training performance}
\label{tab:Train}
  \begin{tabular}{|c|c|c|c|}
    \hline
    Variables & ANN MSE & RF MSE & DT MSE \\
    \hline
    Pressure of the water circuit pump& 1.2e-07 & 1.02e-08 & 7.27e-11\\
    Oil Mist Detector &9.97e-07&9.46e-08&6.39e-10\\
    Fuel Injected & 56 & 6.19 & 0.057\\
    Fuel Pressure & 0.00035 & 2.81e-05 &1.75e-07 \\
    Turbocharger Rotational Speed & 489601 & 50429 & 430 \\
    Rail Fuel Pressure & 27.39 & 2.55 & 0.0107\\
    \hline
  \end{tabular}
\end{table}



\begin{table}[h]
  \centering
\caption{Test performance}
\label{tab:Test}
 \begin{tabular}{|c|c|c|c|}
    \hline
    Variables & ANN MSE & RF MSE & DT MSE \\
    \hline
    Pressure of the water circuit pump& 1.18e-07 & 7.06e-08 & 1.18e-07\\
    Oil Mist Detector &9.88e-07&6.60e-07& 1.10e-06\\
    Fuel Injected & 55 & 42 & 71\\
    Fuel Pressure & 0.00035 & 0.00020 &0.00030 \\
    Turbocharger Rotational Speed & 479274 & 343723 & 574069 \\
    Rail Fuel Pressure & 27.75 & 18.62 & 31.26\\
    \hline
  \end{tabular}
\end{table}


The DT outperforms the other two algorithms during the training phase, but {lacks of generalization capacity} 
during the test phase. The ANN is the worst algorithm. This is probably due to the fact that the ANN is more affected by noise than RF and DT, even after applying the simple {moving} average. RF is the best predictive algorithm because, as an ensemble of several DTs, it implements the best generalization process. Then, the RF has been chosen as the algorithm to test the proposed method for the early detection of catastrophic failures.
Note that the hyper-parameters of the trained ML algorithms are shown in the Tables~\ref{tab:ANNhyp}--\ref{tab:DThyp}.

\begin{table}[ht!]
  \centering
\caption{ANN hyper-parameters}
\label{tab:ANNhyp}
  \begin{tabular}{|c|c|}
    \hline
    Number of layers and neurons & 3 with 32, 24 and 12 neurons\\
    Learning rate& 0.01\\
    Activation function& \textit{ReLu} \\
    Optimizer & SGD\\
    loss & MSE \\
    \hline
  \end{tabular}
\end{table}

\begin{table}[ht!]
  \centering
\caption{RF hyper-parameters}
\label{tab:RFhyp}
  \begin{tabular}{|c|c|}
    \hline
    Number of estimators & 100 \\
    Minimum samples of splits& 2\\
    Minimum samples of leaf& 1 \\
    Maximum number of feature & 1\\
    Criterion & squared error\\
    \hline
  \end{tabular}
\end{table}

\begin{table}[ht!]
  \centering
\caption{DT hyper-parameters}
\label{tab:DThyp}
  \begin{tabular}{|c|c|}
    \hline
    Criterion & squared error\\
    Minimum samples of splits& 2\\
    Minimum samples of leaf& 1 \\
    \hline
  \end{tabular}
\end{table}

{Moreover, Tables~\ref{tab:Train_no_augmentation}--\ref{tab:Test_no_augmentation} show the results of the ML algorithms tuned with the same parameters but without data augmentation. It is clear that we get a loss of performance, so data augmentation is necessary to increase the generalization capability of the chosen algorithms and reduce the training and test errors.

\begin{table}[h]
  \centering
\caption{Training performance without the data augmentation}
\label{tab:Train_no_augmentation}
  \begin{tabular}{|c|c|c|c|}
    \hline
    Variables & ANN MSE & RF MSE & DT MSE \\
    \hline
    Pressure of the water circuit pump& 1.2e-07 & 1.31e-08 & 1.54e-10\\
    Oil Mist Detector &1.52e-06&1.50e-07&1.5e-09\\
    Fuel Injected & 81.6 & 9.8 & 0.117\\
    Fuel Pressure & 0.0005 & 4.85e-05 &4e-07 \\
    Turbocharger Rotational Speed & 649384 & 80000 & 860 \\
    Rail Fuel Pressure & 37.71 & 3.6 & 0.022\\
    \hline
  \end{tabular}
\end{table}

\begin{table}[h]
  \centering
\caption{Test performance without data augmentation}
\label{tab:Test_no_augmentation}
 \begin{tabular}{|c|c|c|c|}
    \hline
    Variables & ANN MSE & RF MSE & DT MSE \\
    \hline
    Pressure of the water circuit pump& 1.22e-07 & 9.61e-08 & 1.59e-07\\
    Oil Mist Detector &1.48e-06&1.063e-06& 1.82e-06\\
    Fuel Injected & 84.7 & 73.30 & 123.96\\
    Fuel Pressure & 0.0053 & 0.0034 &0.00060 \\
    Turbocharger Rotational Speed & 672476 & 595749 & 1000309 \\
    Rail Fuel Pressure & 36.67 & 24.89 & 43\\
    \hline
  \end{tabular}
\end{table}

}

Now, we remind that thresholds are necessary for the early detection of catastrophic failure. While 
traditional warning/alarm signals are provided based on the expert knowledge, no ``limit values" are available for a derivative-based detection logic. Note that the thresholds are usually defined as around 5--10\% higher than the expected values, as shown in \cite{Ceglie, Rubio}. 
Then, we propose to use the maximum value of the first and second derivatives of the 
the deviation that is computed while monitoring a healthy load profile.  

The results on the computation of the {deviation} and its derivatives for the faulty profile 
are shown in Fig.~\ref{fig.derivative_fig1}--\ref{fig.derivative_fig2}. In the figures, the blue curve is the deviation, the orange curve is the first derivative, the green line is the second derivative, the dotted black line is the 5\% traditional warning/alarm signal, and the dotted green and orange lines are the thresholds for the first and second derivative, respectively.  
It can be noted that the derivatives exceed the threshold earlier {than the simple deviation} or even for {the entire duration of} the faulty profile, while the common threshold for the deviation is exceeded later, namely too late for a correct early detection of the catastrophic failure. 

         \begin{figure}[H]
      \centering
      \includegraphics[width=\linewidth]{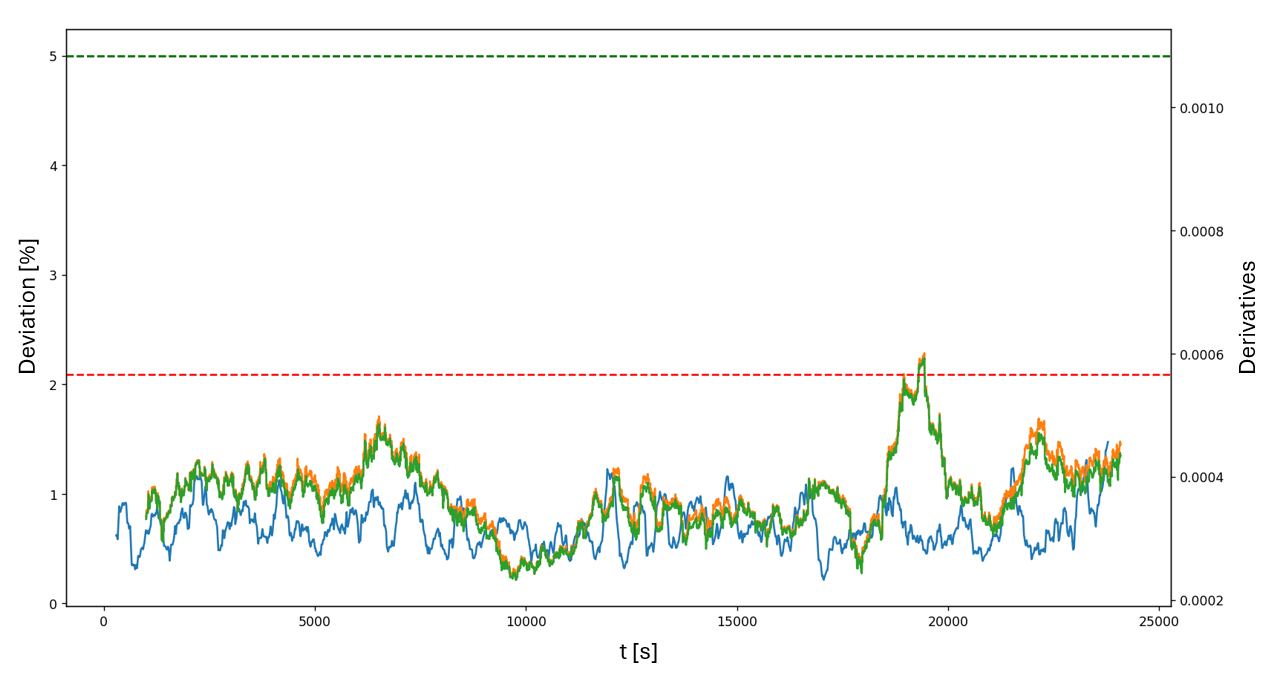}
      \caption{{Deviations and their first and second} derivatives for the fuel pressure. 
      }
      \label{fig.derivative_fig1}
   \end{figure}

           \begin{figure}[H]
      \centering
      \includegraphics[width=\linewidth]{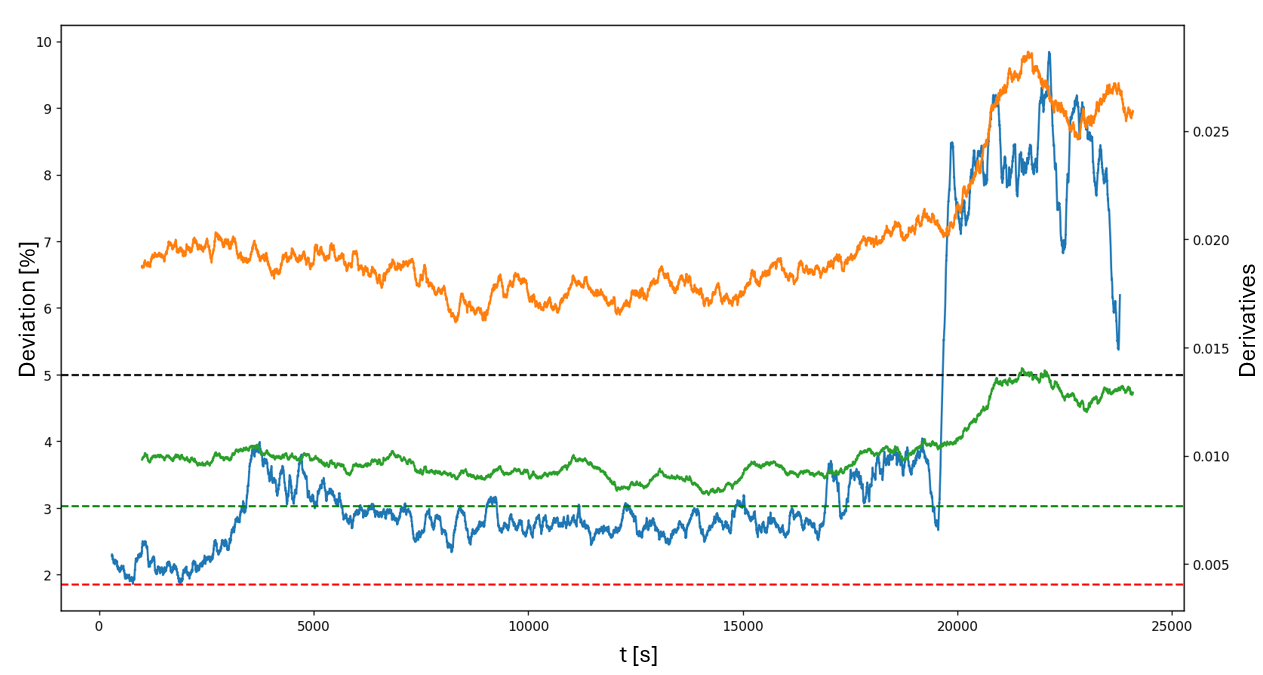}
      \caption{{Deviations and their first and second} derivatives for the {pressure generated by the pump in the water circuit}.  
      }
      \label{fig.derivative_fig2}
   \end{figure}

In summary, if the 
derivatives were checked in parallel to the deviation values, operators would have the opportunity to shut down the engine much earlier than usual and preserve it from a total disruption. Then, by the proposed method, onboard operators would have the opportunity to be warned in time, avoid unexpected failures, then an unexpected loss of power and maneuverability, with the risk of collision and deadly accidents.  
}

\subsection{COMPARATIVE ANALYSIS} \label{comparison}
{
We now compare the proposed algorithm with a common outlier detection ML algorithm, namely the so-called One Class Support Vector Machine (OC-SVM). 
Unlike the traditional SVM, the OC-SVM does not require labeled data, namely data from both the healthy and faulty conditions, but only the healthy ones. In this way, it is trained to recognize if new data belong or not to the nominal behavior. 
{Moreover, it is used in the same condition as the method we propose, that is, without having faulty data at disposal. Namely,} the OC-SVM is a good choice when there is a lack of labeled data or in presence of unbalanced data, as in the maritime sector. Its advantage is that it can correctly {identify} even previously unseen failures; however, the information on the kind of failure affecting the engine is missing. Due to all these reasons, the OC-SVM is the most suitable term of comparison for the considered application.

The OC-SVM works by defining a ``boundary" around the known healthy data in the feature space, which is obtained through a specified kernel function. This boundary is then used to separate healthy from anomaly data.
Then, the OC-SVM is trained on the healthy data by dividing them into a training set with 75\% of the samples and a test set with the remaining 25\% amount of samples. Fig.~\ref{fig.OCSV} reports the OC-SVM outputs on the faulty data. The OC-SVM outputs $-1$ if data are faulty, $+1$ if data are healthy.



           \begin{figure}[H]
      \centering
      \includegraphics[width=\linewidth]{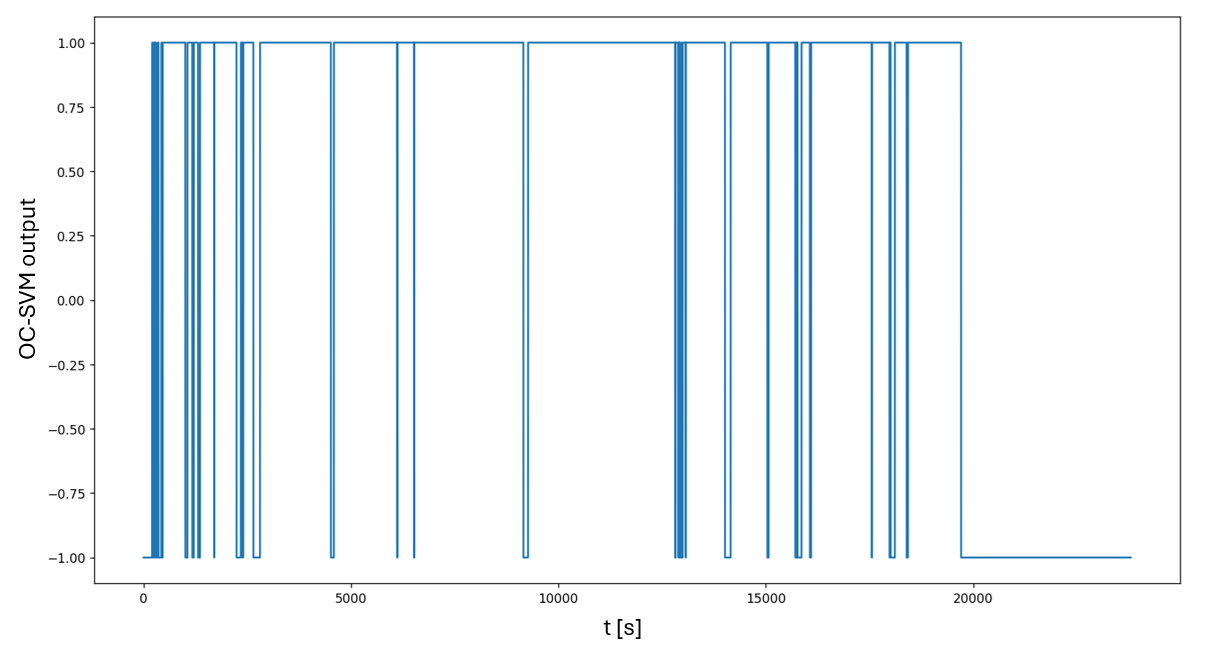}
      \caption{Output of the OC-SVM algorithm.  
      }
      \label{fig.OCSV}
   \end{figure}

It is clear that, at the onset of the failure mode (time $0$ s in Fig.~\ref{fig.OCSV}), the engine's behavior is not sufficiently different from the normal operating mode to allow the OC-SVM to correctly recognize the difference. Namely, the output oscillates between $-1$ and $1$. Since that, the $-1$ predicted before time $20000$ s can be confused with a ``False Negative" (i.e., a false alarm for a fault), and, for that reason, on-board operators can continue to work without shutting down the engine. The fault can be detected only after time $20000$ s (see the constant $-1$ output in Fig.~\ref{fig.OCSV}), when it is too late to shut down and save the engine. On the contrary, checking the derivatives of the {deviation}, especially in the {pressure provided by the pump in the water circuit (see Fig.~\ref{fig.derivative_fig2})}, would allow the ship operators to find out that the engine has an unexpected behavior {and early, namely since time $0$ s in Fig.~\ref{fig.derivative_fig2}.

{Moreover, it is worth noting the effectiveness of the proposed method not only in the case of the pressure generated by the pump in the water circuit, when the derivatives of the deviation are always higher than the established threshold, {but also in case of the fuel pressure, when the derivatives of the deviation exceed the associated threshold for a short time interval only}. 
Namely, Fig. \ref{fig.OCSVvsMethod} shows the comparison between the proposed method and the OC-SVM outputs in case of the fuel pressure. The colored curves are the same used in Fig. \ref{fig.derivative_fig1}, with the addition of the yellow curve for the OC-SVM outputs. Moreover, the blue vertical line indicates the detection time by the proposed method and the vertical violet line the detection time provided by the OC-SVM. Therefore, the proposed method detects the catastrophic failure earlier than the OC-SVM also for this monitored variable, giving the on-shore or onboard operators more time to shut down the engine and thus avoid its total disruption.
}

           \begin{figure}[H]
      \centering
      \includegraphics[width=\linewidth]{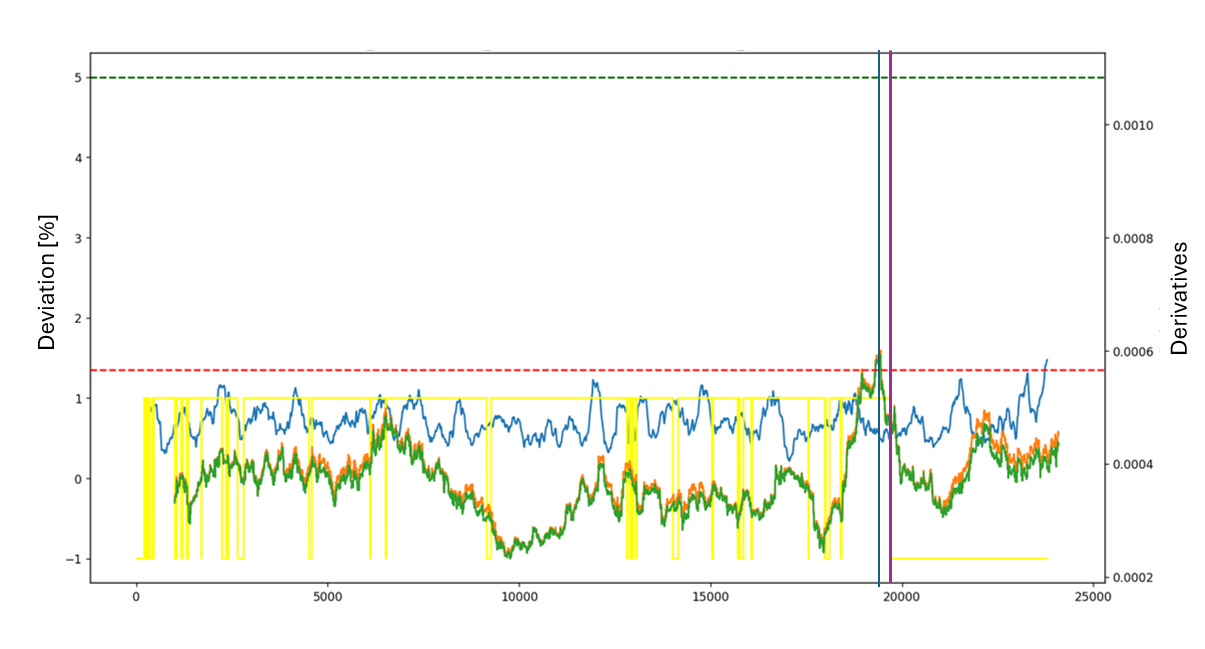}
      \caption{Comparison between the OCSVM outputs and the ones provided by the proposed method for the fuel pressure.  
      }
      \label{fig.OCSVvsMethod}
   \end{figure}

{

Another important observation highlights the robustness of the proposed approach. The high perturbation associated with the catastrophic failure of a collapsed bearing is brought by the necessary replacement, then by the new substitute bearing and consequent slightly different working condition of the engine after maintenance. In this case, the OC-SVM algorithm must be re-trained on new healthy data. Namely, the algorithm bases anomaly detection on all the data acquired in the current condition. Then, a slight change of inputs to the algorithm determines a risk of false failure detection when the engine is simply working in a new, different condition.
On the contrary, the trained RF receives unchanged sensor measurements as inputs, namely the rotational speed and power, which preserve their values in all conditions. Hence, the predictions regard the normal working condition prior to the failure, while the variables measured on the maintained components will change with respect to this previous condition. For that reason, if a mechanical component slightly changes its behavior due to maintenance (e.g., it works at a different temperature), then we may adapt the alarm threshold for the derivatives of behavior deviations to this new normal behavior. 
Thus, we can avoid ``False Positive'' or ``False Negative'' outputs and avoid the re-training of the RF predictive algorithm. However, this advantage is effective only if the new working condition changes few measured variables.

}}


\section{CONCLUSIONS} \label{conclusions}

This paper 
gives 
a contribution 
to the problem of early detection of catastrophic failures in marine Diesel engines.
Classical methods have been developed for traditional failures, namely, the ageing ones. This is why they often fail to detect catastrophic failures on time. On the contrary, we propose a method specifically designed to early detect catastrophic failures, thereby contributing to this important, uncovered topic.
The developed detection logic is based on evaluating the derivatives 
of the difference between the simulated sensor 
measurements and expected values obtained by machine learning algorithms.
Namely, the trend of the derivatives 
is used as an early indicator 
of upcoming, sudden, and severe failures. 
The best results are given by a Random Forest.

It is remarkable that traditional detection methods for marine engines are based on warnings and alarms, which are simply generated
if a
variable under consideration reaches a threshold, 
such that they do not allow us to detect the impending catastrophic failure in time. On the contrary, the proposed method extends the time available 
to perform a rapid intervention and shut down the engine such that its integrity is preserved and the ship with its crew and passengers is protected by an 
accident.  

{
Finally, since this work employed real data of an engine subject to a catastrophic failure, the next step will be to generalize the proposed strategy and verify its robustness 
for different engines and assets.}
{Moreover, 
we will consider the possibility to integrate the proposed strategy into a decision support system.
}

\addtolength{\textheight}{-0cm}   




\section*{ACKNOWLEDGMENT}

Activities described in this paper have been developed within Fincantieri project “Wave 2 the Future” (W2F) funded by the European Union –NextGenerationEU. W2F project is part of IPCEI Hy2Tech.


\end{document}